\documentclass[10pt,twocolumn,letterpaper]{article}

\usepackage{iccv}
\usepackage{times}
\usepackage{epsfig}
\usepackage{graphicx}
\usepackage{amsmath}
\usepackage{amssymb}

\usepackage{scalerel,stackengine}
\stackMath

\renewcommand{\vec}[1]{\boldsymbol{#1}}
\newcommand{\mat}[1]{\mathbf{#1}}
\newcommand{\set}[1]{\mathcal{#1}}

\newcommand{\real}[0]{\mathbb{R}}

\newcommand{\pose}[0]{\vec{\theta}}
\newcommand{\rootor}[0]{\vec{\phi}}

\newcommand{\shape}[0]{\vec{\beta}}

\newcommand{\trans}[0]{\vec{t}}

\newcommand\reallywidehat[1]{%
\savestack{\tmpbox}{\stretchto{%
  \scaleto{%
    \scalerel*[\widthof{\ensuremath{#1}}]{\kern-.6pt\bigwedge\kern-.6pt}%
    {\rule[-\textheight/2]{1ex}{\textheight}}%
  }{\textheight}%
}{0.5ex}}%
\stackon[1pt]{#1}{\tmpbox}%
}

\usepackage{xcolor}
\usepackage{overpic}

\usepackage[pagebackref=true,breaklinks=true,letterpaper=true,colorlinks,bookmarks=false]{hyperref}

\iccvfinalcopy %

\def\iccvPaperID{3848} %

\ificcvfinal\fi

\begin{document}

\title{Generating Continual Human Motion in Diverse 3D Scenes}

\author{Aymen Mir\textsuperscript{1, 2} \qquad Xavier Puig\textsuperscript{3} \qquad Angjoo Kanazawa\textsuperscript{4} \qquad Gerard Pons-Moll\textsuperscript{1, 2}\\\\
{\small \textsuperscript{1} Tübingen AI Center, University of Tübingen, Germany}\\
{\small\textsuperscript{2}Max Planck Institute for Informatics, Saarland Informatics Campus, Germany}\\
{\small\textsuperscript{3}Meta AI Research, }\\
{\small\textsuperscript{4}University of California, Berkeley}\\
{\tt\scriptsize \{aymen.mir,gerard.pons-moll\}@uni-tuebingen.de, xavierpuig@meta.com, kanazawa@eecs.berkeley.edu }}

\makeatletter
\let\@oldmaketitle\@maketitle%
\renewcommand{\@maketitle}{
	\@oldmaketitle%
	\begin{center}
 	\includegraphics[width=1\linewidth]{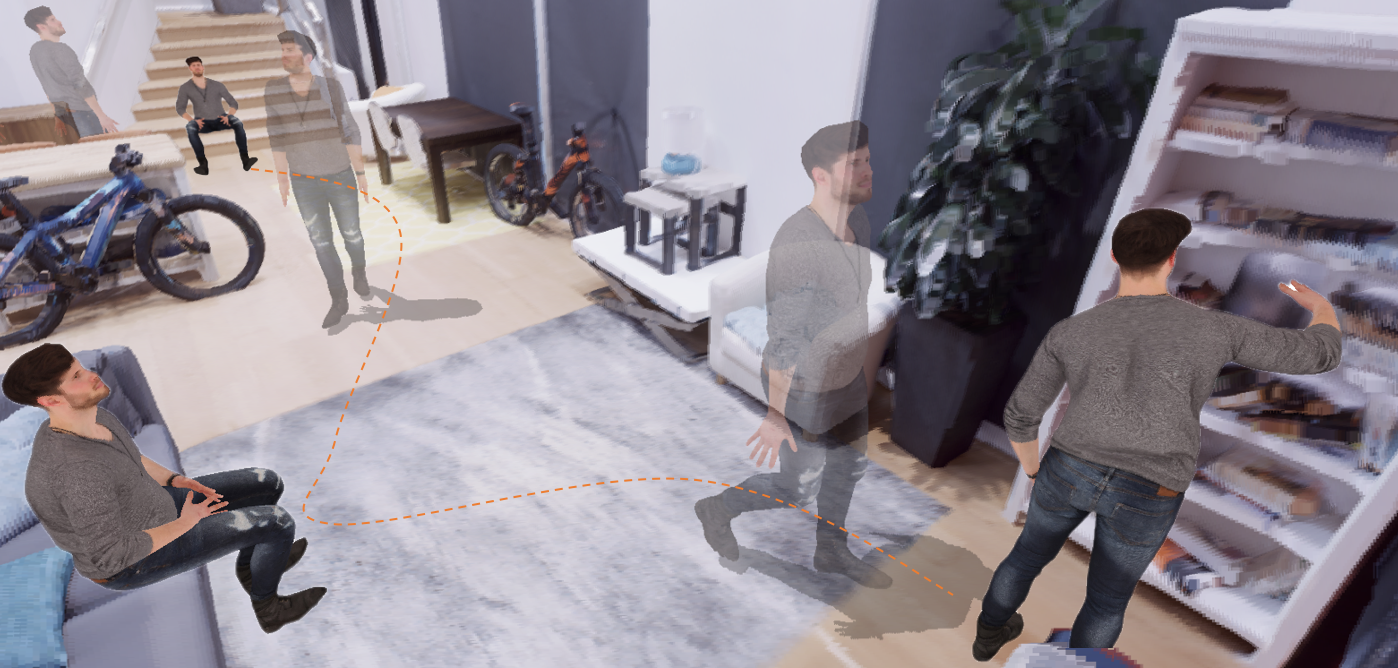}

	\end{center}
    \refstepcounter{figure}\normalfont \footnotesize
Figure~\thefigure. Our method synthesizes diverse animator guided human motion such as sitting and grabbing in diverse 3D scenes. We urge readers to watch the \href{https://www.youtube.com/watch?v=0wZgsdyCT4A&t=1s}{supplementary video} for more results. 

	\label{fig:teaser}
\vspace{1mm}
}
\makeatother
\maketitle
\begin{abstract}
\vspace{-2mm}
We introduce a method to synthesize animator guided human motion across 3D scenes. 
Given a set of sparse (3 or 4) joint locations (such as the location of a person's hand and two feet) and a seed motion sequence in a 3D scene, our method generates a plausible motion sequence starting from the seed motion while satisfying the constraints imposed by the provided keypoints. We decompose the continual motion synthesis problem into walking along paths and transitioning in and out of the actions specified by the keypoints, which enables long generation of motions that satisfy scene constraints without explicitly incorporating scene information. Our method is trained only using scene agnostic mocap data. 
As a result, our approach is deployable across 3D scenes with various geometries. For achieving plausible continual motion synthesis without drift, our key contribution is to generate motion in a goal-centric canonical coordinate frame where the next immediate target is situated at the origin. 
Our model can generate long sequences of diverse actions such as grabbing, sitting and leaning chained together in arbitrary order, demonstrated on scenes of varying geometry: HPS, Replica, Matterport, ScanNet and scenes represented using NeRFs. Several experiments demonstrate that our method outperforms existing methods that navigate paths in 3D scenes. Our webpage is available at \url{https://virtualhumans.mpi-inf.mpg.de/origin_2/}

\end{abstract}
\vspace{-3mm}

\section{Introduction}
\label{sec:introduction}
Our goal is to generate animator guided rich long-term human behavior in arbitrary 3D scenes, including a variety of actions and transitions between them.
Such a system should allow for goal-directed generation of humans moving about from one place to another, for example, walking towards the couch to sit on it, and then standing up and approaching the shelf to grab something from it, as illustrated in Figure~\ref{fig:teaser}. It should allow users to specify with minimal interaction what kind of actions to perform, while keeping the realism and expressivity required for applications such as synthetic data generation, robotics, VR/AR, gaming, etc.

While the community has seen promising progress in animator guided motion synthesis in 3D scenes,
most works are restricted to a single action and do not handle transitions
~\cite{zhang2022couch,wu2022saga,taheri2022goal}, preventing them from producing long range diverse motion. They are also not deployable in a wide variety of real scenes~\cite{starke19neural,wang2021synthesizing, wang2022towards, hassan_samp_2021}.
The reason for this is that they synthesize motion by conditioning on scene geometry and require training on a dataset featuring 3D humans interacting in 3D scenes and objects 
~\cite{hassan2019prox,hassan_samp_2021, zhang2022couch}. Generalizing these methods to arbitrary 3D scenes would require collecting motion data registered to a myriad of possible 3D scenes and objects, which is not scalable.

In contrast, humans can navigate cluttered scenes, pick objects from a shelf they have never seen before, and sit on novel furniture and surfaces. Most of the clutter in the scene is often ignored, and what matters most are not the exact details of the object / scene geometry but whether they afford each action. Our hypothesis is that motion, to a large extent, is driven to avoid obstacles and focused on reaching the next immediate goal or target contacts in the environments. Thus, it should be possible to generate human motion without accounting for all the details in the 3D scene.

Based on this insight, we propose a novel framework for animator-guided motion synthesis in 3D scenes without relying on scene-registered motion data. As such, our method can be trained on regular mocap data, which is relatively easily captured and abundantly available~\cite{mahmood19amass}. 
Since our method does not explicitly condition on the geometry of the scene, it can be deployed across 3D scenes with varied geometry.

Our method relies on two key observations: first, we can represent actions in a 3D scene as a set of sparse desired target contacts (we use 3 or 4 contacts such as the location of the two feet and a hand or the location of two feet and the root) to be reached, which we refer to as \emph{action keypoints}. These keypoints can be provided by an animator using an interface or generated by automated heuristics, allowing animators to trade off the speed and control over the generation motion. An interesting finding in this paper is that \emph{action keypoints} are a powerful abstraction of several actions in 3D scenes, and can be used to execute instructions such as ``sit there" or ``grab at this height".  %
Second, avoiding obstacles in 3D scenes can be achieved by path following. The challenge is to follow arbitrarily long paths, smoothly making the human transition into and out of the action, and then walk towards the next target. For this, we break down motion into three pieces: walking, transition into and out of an action. For path following and transitions, we introduce the idea of training a motion synthesis model entirely with \emph{scene-agnostic motion data} to reach the origin of a \emph{canonical coordinate frame}. For navigating paths, this model is sampled iteratively to converge at the origin of the \emph{canonical coordinate frame} defined using waypoints and tangents on the path. For transitions in and out of actions, motion is synthesized by placing target poses at the origin of the canonical coordinate frame. By iteratively synthesizing motion in the \emph{canonical coordinate frame}, our method allows for long range motion synthesis that transitions between walks and various actions in a 3D scene.

Unlike existing methods for motion synthesis \cite{hassan_samp_2021, starke19neural}, our method allows for synthesizing motion without requiring any manual phase or action annotation.

For the first time, we demonstrate long-range human motion synthesis on a wide range of scene datasets: Replica~\cite{replica19arxiv}, Matterport~\cite{niessner2017Matterport3D}, HPS~\cite{mir20hps}, Scannet~\cite{dai2017scannet} and a NeRF scene. Furthermore, we show that our model can perform actions at different places, such as grabbing from any shelf, table or cabinet at any height or sitting on any surface that affords sitting. We will make our code and models publicly available which can be used by animators to synthesize goal directed human motion across 3D scenes.

To summarize, our contributions are as follows:
\begin{itemize}
\item We present a method that departs from existing methods for motion synthesis in 3D scenes by only using regular motion capture data and that is deployable across varied 3D scenes.

\item We introduce a novel idea of iteratively converging motion at the origin of a canonical coordinate frame, which allows to synthesize long-range motion in 3D scenes.

\end{itemize}

\section{Related Work}
\label{sec:related}
\begin{figure*}[t]
    \centering
    \includegraphics[width=\linewidth]{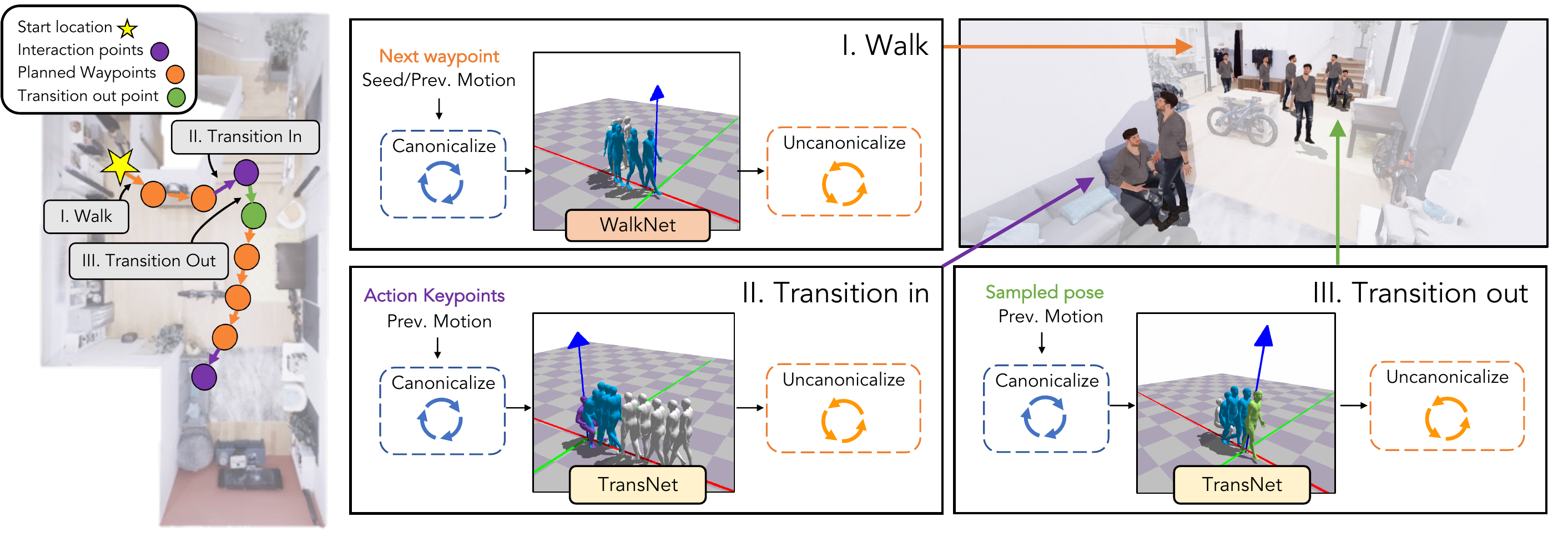}
    \caption{ Overview of our method. We generate human motion satisfying keypoint constraints by diving it into 3 stages: a \textit{Walk Motion}, which animates the human as it walks between keypoints, a \textit{Transition-In}, which blends the walking motion with the pose specified by the keypoints and a \textit{Transition-Out}, which animates the human back to the walking pose. We use an autoregressive transformer, \textit{WalkNet}, to synthesize the walking motion, and a masked-autoencoder transformer to generate the blending motion. By moving the motion into a Goal-Centric Canonical Coordinate Frame our method can generalize to a wide set of 3D scenes.}
    \label{fig:system}
\end{figure*}
\paragraph{Human Motion Prediction without the 3D scene.}
Predicting the dynamics of human motion is a long studied problem in computer vision and graphics. Classic works explored using Hidden Markov Chains \cite{brand2000markov} and Gaussian Processes \cite{wang2007gaussian}, physics based models~\cite{liu2005learning} for predicting future motion. Recently, recurrent neural networks \cite{graves2013generating,lstm97hoch} have been used for motion prediction \cite{fragkiadaki2015recurrent, martinez2017human, alahi16sociallstm} also in combination with Graph Neural Networks \cite{kipf15gcn, mao2019learning, Li_2020_CVPR, Dang:ICCV2021}, and variational Auto-encoders \cite{kingma2013auto} to add diversity \cite{habibie17recurrent, Zhang:CVPR:2021}. Yuan et al. \cite{yuan2020dlow}. An intrinsic problem of recurrent methods is that they drift over time~\cite{aksan2020spatiotemporal}. 

More recent approaches employ transformers to generate unconditional or text and music conditioned motion sequences~\cite{aksan2020spatiotemporal,li2021learn, li2020learning, petrovich21actor, petrovich22temos}.
We also build on transformer architectures but aim to generate motion in real 3D scenes. %

Motion Inbetweening \cite{duan2021singleshot, harvey2020robust, oreshkin2022motion, yuan2022glamr, Aksan_2019_ICCV, kaufmann20convolutional} is another classic paradigm for motion synthesis where the task is to fill in frames between animator provided keyframes.

Our approach builds on recent progress in transformer architectures~\cite{li2021learn}, and classical ideas such as motion inbetweening, combined with the novel idea of a canonical coordinate frame and action keypoint representation in order to generate motion in 3D scenes. 

\paragraph{Character Control in Video Games.}
Motion matching~\cite{reitsma2007evaluating}, its learnt-variant \cite{clavet16motionmatching, holden2020learned} and motion graphs ~\cite{lee2002interactive,fang2003efficient,kovar2008motion,safonova2004synthesizing,Safonova:2007:InterpolatedGraphs} are classical methods often employed in the industry for generating kinematic motion sequences, controlled by environment and user specified constraints. Similar to our goal, some works~\cite{puig2018virtualhome, caoHMP2020} use a combination of these approaches and IK to generate human behaviors in synthetic scenes. However, these approaches require significant human effort to author realistic animations, and IK approaches easily produce non-realistic animations.    

Deep learning variants such as Holden~\etal \cite{holden2017phase} introduce phase-conditioning in a RNN to model the periodic nature of walking motion. %
In several works by Starke~\etal \cite{starke19neural, starke21martialarts, starke20local} the idea of local phases is extended to synthesize scene aware motion, basketball motion and martial arts motion. All these methods generate convincing motion but phases are non-intuitive for non-periodic motion and often require manual labelling.

\paragraph{Static Human Pose Conditioned on Scenes.}
The relationship between humans, scenes, and objects is another recurrent subject of study in computer vision and graphics. Classical works include methods for 3D object detection~\cite{gupta2007objects, gupta20113d} and affordance prediction using human poses~\cite{delaitre12scenesemantics, grabner2011makes, fouhey2014people}.

Several recent works, generate plausible static poses conditioned on the a 3D scene~\cite{li2019puttingnvidia, Zhang:CVPR:2021, wang2017binge, zhang20place, hassan21cvpr, Zhao_ECCV2022} using recently captured human interaction datasets
~\cite{hassan2019prox, mir20hps, savva2016pigraphs, bhatnagar22behave, taheri20grab, cao2020long}. Instead of static poses, we generate \emph{motion} coherent with the scene which is significantly harder.

\paragraph{Scene Aware Motion Synthesis.}
Some works leverage reinforcement learning to synthesize navigation in \emph{synthetic} 3D scenes~\cite{ling20motionvae,zhang2022wanderings}. Other works focus on a single action, such as grabbing \cite{taheri2022goal, wu2022saga} but do not demonstrate transitions to new motions. These methods are not demonstrated in real 3D scenes with multiple objects and clutter. 
Recent real interaction datasets~\cite{hassan2019prox, mir20hps, savva2016pigraphs, bhatnagar22behave, taheri20grab, cao2020long} have powered methods to synthesize 3D scene aware motion \cite{wang2022towards, wang2021synthesizing,cao2020long,wang2021scene}. 
These datasets are crucial to drive progress, but do not capture the richness and variety of real world scenes. Hence, these methods are often demonstrated only on small scenes from PROX \cite{hassan2019prox} and Matterport \cite{niessner2017Matterport3D}.

\begin{figure*}[t]
    \centering
    \begin{overpic}[width=\linewidth]{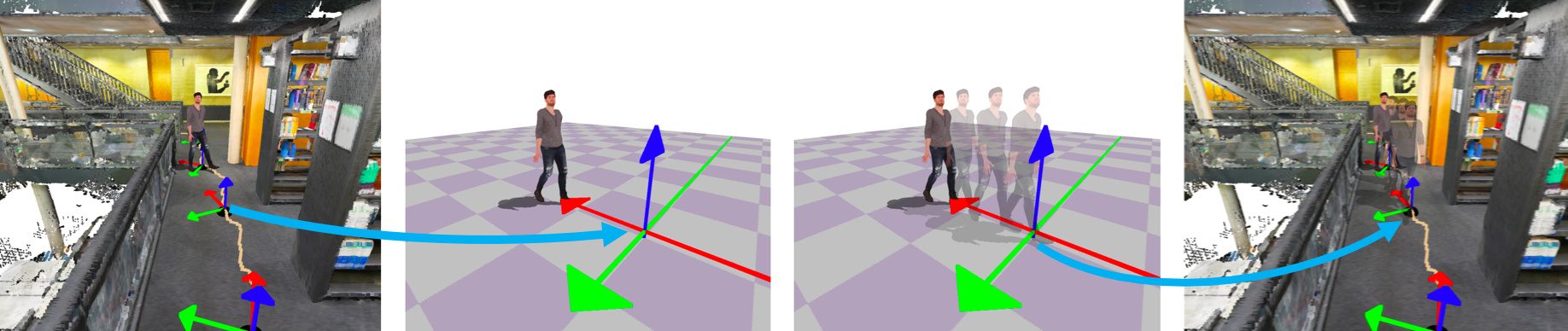}
    \put(12, -1.5){a)}
    \put(38, -1.5){b)}
    \put(62, -1.5){c)}
    \put(88, -1.5){d)}
    \end{overpic}
\vspace*{0.1mm}
\caption{a) Using keypoints and tangents along a path, we move motion from the scene coordinate frame into b) the goal-centric canonical coordinate frame, where c) \textit{WalkNet} synthesizes motion that converges at the origin of the coordinate frame. d) Once the synthesized motion reaches the origin, we move it back to the scene coordinate frame. }
\label{fig:walking}
\end{figure*}

We draw inspiration from Hassan et al. \cite{hassan_samp_2021} which combine path planning with neural motion synthesis, and from Zhang et al. \cite{zhang2022couch}  
which synthesize contact controlled human chair interaction. 
These methods require the geometry of the isolated interacting object as input, which make them hard to generalize to real 3D scenes. 
Unlike these methods, we demonstrate \emph{long chained sequences of actions} in \emph{generic real 3D scenes}, which is enabled with our origin canonicalization and action keypoints.

\section{Method}
\label{sec:method}
Our method takes as input a seed motion sequence and a list of action keypoints $\{ \mathbf{a}_1,\hdots,\mathbf{a}_n\}$ specifying interactions at different locations in the scene. These keypoints can be specified by users or generated using language commands and scene segmentations (Sec. \ref{sec:kps}).
Our goal is to synthesize motion that starts at the seed motion and transitions in and out each of the action keypoints in the input list. 

The first step is to optimize for a pose that fits the action keypoints at target locations using Inverse Kinematics and a pose-prior~\ref{sec:ik}). These poses along with the starting seed motion act as anchors to guide the motion synthesis process.

Using scene-agnostic motion capture data placed in a goal-centric canonical coordinate frame (Sec. \ref{sec:goal_canon}), we train \textit{Walknet} (Sec. \ref{sec:walk}) to synthesize walking motion that converges at the origin of a canonical coordinate frame, and \textit{TransNet} (Sec \ref{sec:transition}) that synthesizes motion inbetween a seed motion sequence and a target pose also at the origin.
At test time (see Fig. \ref{fig:system}), \textit{WalkNet} is used to reach canonicalized intermediate goals along a path computed with a path planning algorithm, thus creating long motion by successively reaching the origin. Once the walking motion reaches the vicinity of an anchor pose, \textit{TransNet} synthesizes transition from walking motion to the anchor pose and vice versa. This allows to synthesize motion in 3D scenes without the need for motion data coupled with 3D scenes. Our framework is general and highly modular, which allows it to be updated with novel methods for motion synthesis.

\subsection{SMPL Body Model}
\label{subsec:meth_smpl}
We use the SMPL body model  \cite{smpl2015loper} to represent the human subject. SMPL is a differentiable function $M(\rootor, \pose, \trans, \shape) $ that maps global body orientation $\rootor$, pose $\pose$, translation $\trans$ and shape $\shape$ parameters to the vertices of a human mesh along with the 3D joint locations of the SMPL skeleton. 
We assume that $\shape$ remains static throughout our method. We denote motion sequences as an ordered list of SMPL parameter tuples. For example $\set{C} = [(\vec{r}, \rootor, \pose)_j]_{j = 1:D}$ denotes a motion sequence of $D$ frames.  %

\subsection{Generating Keypoints in a Scene}
\label{sec:kps}

\begin{figure}
\includegraphics[width=\linewidth]{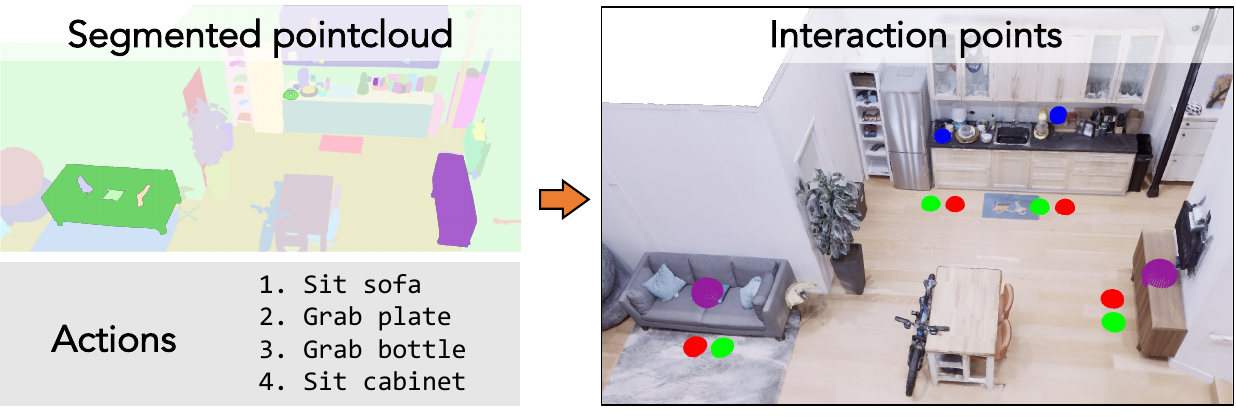}
\caption{Using language instruction and semantic segmentation, keypoints can be automatically placed in a 3D scene.}
\label{fig:kps}
\end{figure}

Keypoints can be efficiently collected using a 3D user interface, as described in the supp. mat, or keypoints can be inferred from the geometry of the scene, and can be therefore generated via action labels or language. An example of automatic KP generation can be seen in Fig.~\ref{fig:kps}. Given a point cloud of the scene with semantic labels and a language description of a task, we can use simple heuristics to generate keypoints that can synthesize the described motion. More details can be found in the supp. mat.         

\subsection{From Action Keypoints to an Anchor Pose}

\label{sec:ik}

The first step is to infer a pose from the action keypoints in a target location $\mathbf{a}=\{\vec{k}_i\}_{i=1}^P$, where $\vec{k}_i \in \mathbb{R}^3$ indicates the desired locations for corresponding SMPL joints denoted as $m_{i}(\cdot)$. We find as few as three to four joints ($P=3,4$) are usually sufficient. Since the problem is heavily under-constrained we optimize the latent space of VPOSER~\cite{SMPL-X:2019} dennoted as $\mathbf{z}$. Denoting $f(\mathbf{z})\mapsto (\phi,\theta)$
as the mapping from the latent space $\mathbf{z}$ to the SMPL pose parameters, we minimize the following objective
\begin{equation}
    \label{eq:vposer}
    \vec{z}, \vec{t} = \arg \min_{\vec{z}, \trans} \sum_{i=1}^{P}||m_i(f(\vec{z}), \trans) - \mathbf{k}_i||_2
\end{equation}
Please see the supplementary material for further details to make the optimization well behaved.  
We repeat this step for each target action $\mathbf{a}_1\hdots \mathbf{a}_N$, obtaining $N$ pose-anchors $\set{A} = \{ \trans^{A}_i, \rootor^{A}_i, \pose^{A}_i \}_{i=1:N}$. 

\subsection{Canonical Coordinate Frame}
\begin{figure*}
    \centering
    \begin{overpic}[width=\linewidth]{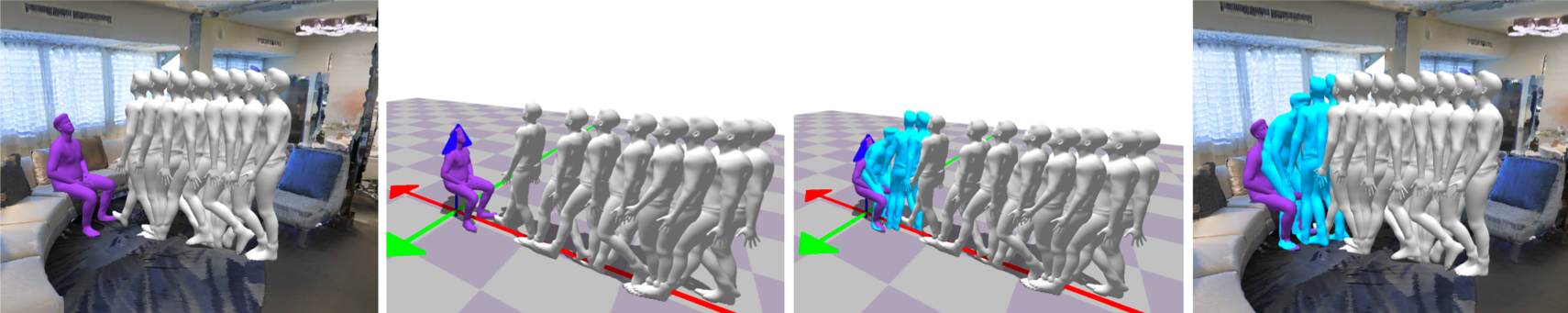}
    \put(12, -1.5){a)}
    \put(38, -1.5){b)}
    \put(62, -1.5){c)}
    \put(88, -1.5){d)}
    \end{overpic}
\vspace*{0.1mm}
\caption{
Using a) the motion-anchor pose in the 3D scene (purple), b) we move the motion sequence into the canonical coordinate frame. c) There \textit{TransNet} synthesizes transitions (blue) between the input motion and the pose placed at the origin (purple). d) Once the motion is synthesized, we move it back to the scene coordinate frame.}
\label{fig:transitions}
\end{figure*}
\label{sec:goal_canon}

One of our key ideas to sythesize motion in 3D scenes is to make transformers synthesize motion that always converge at the origin of a canonical coordinate frame. This way at test time long motion is composed by consecutively going to the next goal placed at the origin. 
Thus, we canonicalize the training sequence clips by using the planar translation $\vec{t}_C$, and rotation $\mat{R_C}$ of the last frame in a sequence clip as follows

\begin{equation}
\rootor^{C}_j = \mat{R}^{-1}_C \rootor_j \enspace , 
\vec{r}^{C}_j  = \mat{R}^{-1}_C (\vec{r}_j  - \vec{t}_C  )\enspace .
\label{eq:canonicalize}
\end{equation}

By construction, this transformation outputs a new set of $L$ frames  $[(\vec{r}^{C}, \rootor^{C}, \pose)_j]_{j=1:L}$, where the last pose is at the origin and oriented towards a canonical axis of orientation $\gamma$ (arbitrary fixed axis).
Let $\mat{X}$ denote a matrix whose columns are vectorized motion parameters (pose and translation combined) 
We will use the following notation to denote the canonicalization in Eq.~\eqref{eq:canonicalize} for a full sequence as
\begin{equation}
    \label{eq:canon_final}
    \mat{X}^C = C (\mat{X};\mat{R}_C,\mat{t}_C)
\end{equation}

\looseness=-1 Synthesizing motion in the goal-centric canonical coordinate frame, allows us to synthesize walking motion along paths in a 3D scene (Sec. \ref{sec:walk}) and transitions in and out of actions (Sec. \ref{sec:transition}) without the need for scene registered data.

\subsection{WalkNet}
\label{sec:walk}
\paragraph{Training.} Using walking sequence clips of variable lenght $L$ canonicalized (last pose at origin), we train \textit{WalkNet}. 
\textit{WalkNet} takes $K$  motion frames as input $\set{W}_{inp}= [ (\vec{r}^W, \rootor^W, \pose^W )_j]_{j=1:K}$ and predicts the next $K$ frames in the sequence  $\set{W}_{out} = [ (\vec{r}^W, \rootor^W, \pose^W)_j]_{j=K:2K}$.
The training sub-clips of size $2K<L$ are randomly sampled from the training walking sequences. 

Expressing sequences as matrices (columns are translations and poses) as explained in the previous section, the transformer takes as input a matrix $\mat{X}_{in} \in \real^{K \times 219}$
and outputs a matrix $\mat{X}_{out}\in \real^{K \times 219}$. 
We denote the learned mapping as $T:\mathbf{X}_{in} \mapsto \mathbf{X}_{out}$.
Note that we input the pose as vectorized joint rotation matrices, which make learning more stable compared to using joint angles.

\paragraph{Test time.} \looseness=-1
We use the \textit{WalkNet} to follow long paths, by breaking the path into intermediate goals that are canonicalized to the origin (Fig. \ref{fig:walking}). To traverse scenes avoiding obstacles, we compute the path 
between the seed motion $\set{I}$ and the first anchor pose $\set{A}_1$ using A-star\cite{Hart1968}. 
Along the path, we sample $P$ goals and compute tangents to the path: $ \{\vec{q}_{p},  \vec{l}_{p} \in \real^3 \}_{p=1 \dots P}$. Then we recursively canonicalize such that tangents $\mat{l}_p$ align with the canonical axis $\gamma$. Hence, canonical translation and rotation are computed as follows
\begin{equation}
\label{eq:matrix_com}
\mat{t}_C=\vec{q}_p, \qquad \mat{R}_C = \mathrm{exp}(\reallywidehat{\vec{l}_{p} \times \vec{\gamma}}) 
\end{equation}
where $\exp(\cdot)$ is the exponential map recovering the rotation from the screw-symmetric matrix $\reallywidehat{\vec{l}_{p} \times \vec{\gamma}}$.
With this, the motion sequence from goal $p-1$ to goal $p$ is obtained by canonicalizing, predicting future motion with the learned mapping $T$ and uncanonicalizing
\begin{equation}
    \mat{X}_{in} \xrightarrow[]{C(\cdot,\mat{R}_C,\mat{t}_C)} \mat{X}_{in}^C \xrightarrow[]{T} \mat{X}_{out}^C\xrightarrow[]{C(\cdot,\mat{R}_C^T,-\mat{t}_C)} \mat{X}_{out}.
\end{equation}
Although the transformer outputs K future frames, at test time, we use it recursively with a stride of 1 for better performance. That means we effectively predict one pose at a time, and we discard the $K+1:2K$ frames. 
In this manner, the motion always goes to the origin, we never have to explicitly send the goal coordinates as input to the network, and we do not drift. When we are sufficiently close to an anchor pose, we predict the transition with TransNet.

\subsection{TransNet}
\label{sec:transition}
We synthesize transitions between walks and actions again in a canonicalized frame. 
To do so, we train \textit{TransNet} - a transformer based motion inbetweener - using AMASS sequences placed in the canonical coordinate frame. 
The task of \textit{TransNet} is to fill in the motion from a seed sequence $\mat{X}_{in}$ to a target \emph{anchor pose}.

\paragraph{Training.}
We train \textit{TransNet} by asking it to recover training clips from masked out ones. We observe that directly asking to infill many frames does not work reliably. Inspired by training of language models \cite{bert}, we progressively grow the mask during training until the desired length. Formally, let $\mat{X}$ be a training clip of length $M$, let $\mat{V} \in [0,1]^{M \times 219}$ be a matrix mask with zero-column vectors for the frames that need to be infilled. The network is tasked to recover $\mat{X}$ from the masked out matrix  $\mat{X} \odot \mat{V}$. The mask $\mat{V}$ is progressively grown to mask all the motion frames between $\frac{M}{2}$ to $M-1$ frames -- everything except the seed motion and the last anchor pose. For more details, please see supp. mat.

\textbf{Test time}
We use \textit{TransNet} to synthesize transitions in 3D scenes by moving $\frac{M}{2}$ frames of a motion sequence into the canonical coordinate frame by using the orientation and position of the motion-anchor pose - the motion-anchor pose is thus placed at the origin of the canonical coordinate frame. \textit{TransNet} is then tasked to infill the missing frames.  (Fig.~\ref{fig:transitions})

\subsection{Chained actions}
With our models and representations we can chain actions trivially. 
At run time, we have to satisfy an arbitrary number of actions keypoints $\{\vec{a}_1,\hdots \vec{a}_N\}$ at different locations. First we compute anchor poses as explained in Sec.~\ref{sec:ik}. Obstacle free paths connecting the locations of actions are computed with A*.  
We rely on \textit{WalkNet} to follow paths until we are sufficiently close to the first anchor pose. Feeding \text{TransNet} with the last $M/2$ predicted frames of \textit{WalkNet} and the anchor pose, we predict the transition into the first anchor pose. To transition out we also use \text{TransNet} with no modification. We sample a location along the path from $\vec{a}_1$ to $\vec{a}_2$ at a fixed distance $\delta$ and place a walking pose from our database. \text{TransNet} then can transition into this walking pose (Fig. \ref{fig:transitions}). Then we activate \textit{WalkNet} and the process is repeated until all actions are executed. 
In addition, we can repeatedly use \text{TransNet} to execute several actions at the same location, like grabing at different heights.

\section{Experiments}
\label{sec:experiments}
In this section we present implementation details of our method. Next, we compare our approach with existing methods. Our experiments show that we clearly outperform existing baselines. Next, we ablate our design choices and finally present qualitative results of our method. 

\subsection{Implementation Details}
\textit{WalkNet} and \textit{TransNet} are BERT \cite{bert} style full-attention transformers. Both consist of 3 attention layers - each composed of 8 attention heads. We use an embedding size of 512 for both transformers. For more details please see the supplementary material. For training both transformers, we set the learning rate to $1e^{-5}$. Both networks are trained using an $L2$ loss. We set $M=120$ and $K=30$. We experimented with three different values of $M$ and found that $M=120$ produces the least foot-skating. Please see the supplementary material for this experiments. 
\begin{table*}[t]
    \centering
    \begin{tabular}{l|c c|c c|c c}
    \hline
          & Ours & SAMP  & Ours & GAMMA & Ours & Wang et al.  \\
    \hline
         Which motion is most realistic (\%) $\uparrow$ & \textbf{71.8} & 28.2  &  \textbf{95.6}&  4.4 &  \textbf{100} & 0 \\
         Which motion satisfies scene constraints best (\%) $\uparrow$ & \textbf{76.8} & 23.2  &  \textbf{100}&  0 &  \textbf{100} & 0 \\
    \hline
    \end{tabular}
    \caption{Comparisons between our method and existing baselines using a perceptual study.}
    \label{tab:user_study}
\end{table*}

\begin{table*}[]
    \begin{center}
{
    \begin{tabular}{l|c c|c c|c c}
        \hline
          & Language & Manual  & \textit{WalkNet} & MoGlow & \textit{TransNet} & NeMF \\
        \hline

         Foot Skate (cm/f) $\downarrow$ & 0.93 & \textbf{0.92}  & \textbf{0.91} & 1.88 & \textbf{1.1} & 1.54\\
         User Study (\%) $\uparrow$ & \textbf{53.8} & 46.2 & \textbf{75.7} & 24.3 & \textbf{66.8} & 33.2\\
         \hline

    \end{tabular}
    }
    \end{center}
    \vspace{-3mm}
    \caption{ Analysis of different components in our method. We compare our method with different baselines across three design components: using language based or manually specified keypoints, the walking motion and the transition motion. }
    \label{tab:table}
\end{table*}

\begin{table}[t]
    \centering
    \begin{tabular}{l c c c c}
    \hline
         & Ours & SAMP  & GAMMA & Wang et al. \\
    \hline
          Foot-skate $\downarrow$& \textbf{0.91} & 1.34 & 0.94 & 4.53 \\
    \hline

    \end{tabular}
    \caption{Comparisons between our method and existing baselines using the foot-skate metric.}
    \label{tab:foot_skate}
    \vspace{-5mm}
\end{table}

\subsection{Datasets}
\textbf{Motion Data:} To train \textit{TransNet} and \textit{Walknet} we use the large mocap dataset \textbf{AMASS} \cite{mahmood19amass}. For exact details how this is done, please see the supplementary material.

\textbf{Scene Datasets:} We demonstrate that our method is able to generate realistic human motion in scenes from \textbf{Matterport3D}, \textbf{HPS}, \textbf{Replica} and
\textbf{ScanNet} datasets. All these datasets have been reconstructed using RGB-D scanners or LIDAR scanners and contain scans with sizes ranging from 20 $m^2$ to 1000 $m^2$. While Replica, Matterport scenes contain perfect geometry, ScanNet scenes do not. Our method is able to generalize across all these scenes.

\subsection{Evaluation Metrics: }
\label{subsec:eval}

We compare our method with existing baselines using perceptual studies and a foot skate metric. Additionally, we ablate various components of our method with the same foot skate metric.

\textbf{Perceptual Study: } We synthesize two motion sequences - one using our method and another using a baseline method and show the two synthesized sequences to participants in our perceptual study. The participant is asked to answer ``Which motion looks most realistic?" and ``Which motion satisfies scene constraints best?". The study is conducted in such a manner that the participant is forced to choose one of two motions in front of him.

\textbf{Foot Skating (FS): } 
The foot-skate metric measures how much foot-skate occurs during a synthesized motion measured in cm/frame. For N frames, it is defined as:
$$
s = \sum_{p = 1}^{N}[v_p(2 - 2^{\frac{h_p}{H}}) \mathbf{1}_{h_p <= H}]
$$
where $h_p$ is the height of the vertex and $v_p$ is the velocity of a foot vertex on the right toe in frame $p$ and $H = 2.5$ cm

\subsection{Comparison with Baselines}
\label{subsec:baseline}
As aforementioned, no method addresses the task of continual motion synthesis in arbitrary 3D scenes. For completeness we do our best to compare our approach with three existing methods: SAMP \cite{hassan_samp_2021}, GAMMA \cite{zhang2022wanderings}, Wang et al. \cite{wang2022towards} which all generate animator guided motion by navigating A* paths in 3D scenes. Though, these methods use different forms of animator guidance - such as action labels, we modify them by incorporating the KP information used by our method. Note that except GAMMA, none of these baselines can be deployed in arbitrary 3D scenes without significant modifications, as described below.

\textbf{SAMP:} SAMP is written entirely in Unity and can only synthesize sitting and lying actions in synthetic scenes. Unlike SAMP our method requires no manual action annotation. The object of interaction and the action to perform are the animator guidance provided as input to SAMP. SAMP synthesizes motion by explicitly conditioning on the geometry of the object of interaction, and by navigating A* paths. 
For comparison with SAMP, we represent the object of interaction in one of our test scenes with a synthetic object in Unity. Using KPs, we represent the orientation of action in our test scene and use this orientation to port A* paths used in our test 3D scene into Unity and run the publicly available code of SAMP. For exact details, please see the supp. mat.
Note that SAMP cannot synthesize chained actions nor can it be deployed in arbitrary 3D scenes. For instance it cannot sit on stairs nor can it perform a grabbing action near a bookshelf. The comparison is included for completeness as SAMP also navigates A* paths. 

\textbf{Wang et al.}: We run the pre-trained code of Wang et al. on scenes from HPS, Replica and Matterport Datasets. Instead of using action labels to generate anchor poses as done in the original paper, we replace this step with the motion anchors generated using our inverse kinematics step. Since Wang et al. \cite{wang2022towards} is trained using the PROX dataset and synthesizes navigational motion across A* paths by explicitly conditioning on scene geometry, it does not generalize at all to 3D scenes beyond these datasets. 

\textbf{GAMMA:} 
GAMMA only navigates 3D scenes and is unable to synthesize human-scene interaction. Similiar to the navigation part of our method, it uses the start and end of a path as animator guidance. For the purpose of this comparison, we generate a set of paths in 3D scenes using A* and synthesize walking motion along this path using GAMMA and our method. GAMMA is unable to follow the exact waypoints of the path and as such produces significant interpenetrations with the 3D scene. 

For visualizations of motion synthesized by these baselines, please see the supplementary video.
We synthesize 5 motion sequences of a total duration of 300 seconds using each method in 5 different scenes for our perceptual study. In Tab. \ref{tab:user_study}, we report the results of our perceptual study with 50 participants (see Sec. \ref{subsec:eval}). Each column corresponds to the percentage of users who choose the method corresponding to the column heading. Our results are preferred by a vast majority of the participants. In Tab. \ref{tab:foot_skate}, we report the numbers corresponding to foot skate metric.

\subsection{Ablation Studies}
\vspace{-1mm}
\label{subsec:abl}
\textbf{Can \textit{TransNet} be replaced with other inbetweeners?}
We compare \textit{TransNet} with the SoTA inbetweening method NeMF \cite{he2022nemf} for the task of transitioning in and out of actions. For our task of infilling $\frac{M}{2} - 1$ frames in the canonical coordinate frame, 
\textit{TransNet} produces more natural motion and less foot skating. We hypothesize that this occurs as NeMF is a general purpose inbetweener that can infill an arbritrary number of frames, whereas \textit{TransNet} is a motion inbetweener custom designed for the purpose of infilling $\frac{M}{2} - 1$ motion frames in the canonical coordinate frame. 
We conduct a new user study with 36 participants, asking users to rate the naturalness of 20 motion sequences by NeMF and \textit{TransNet}. Results are reported in Tab \ref{tab:table}.

\begin{center}
\begin{figure*}[t!]
\includegraphics[width=\linewidth]{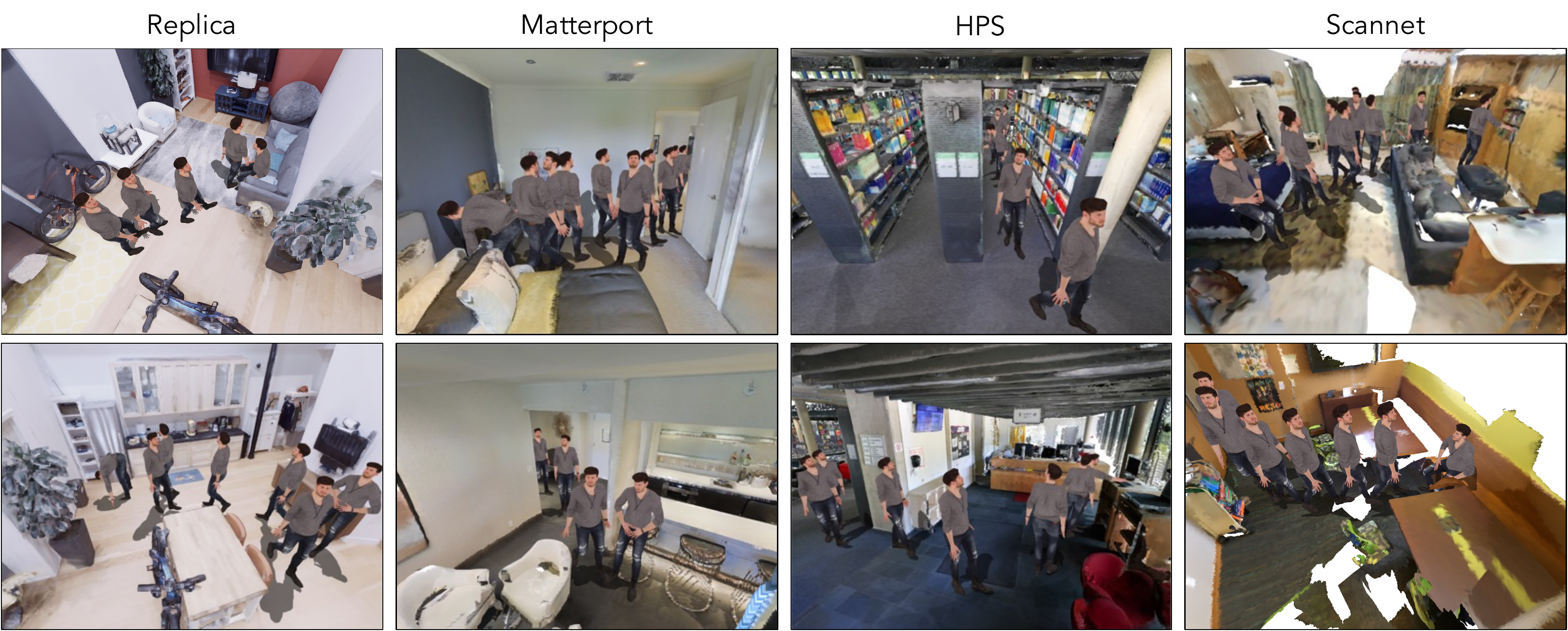}

\caption{Our method allows to generate motion that generalizes across different scenes. Here we show motion generation in scenes from 4 different datasets: Replica~\cite{replica19arxiv}, Matterport~\cite{niessner2017Matterport3D}, HPS~\cite{mir20hps} and Scannet~\cite{dai2017scannet}. }
\label{fig:qual_scene}
\end{figure*}
\end{center}

\begin{center}
\begin{figure*}[t!]
\includegraphics[width=\linewidth]{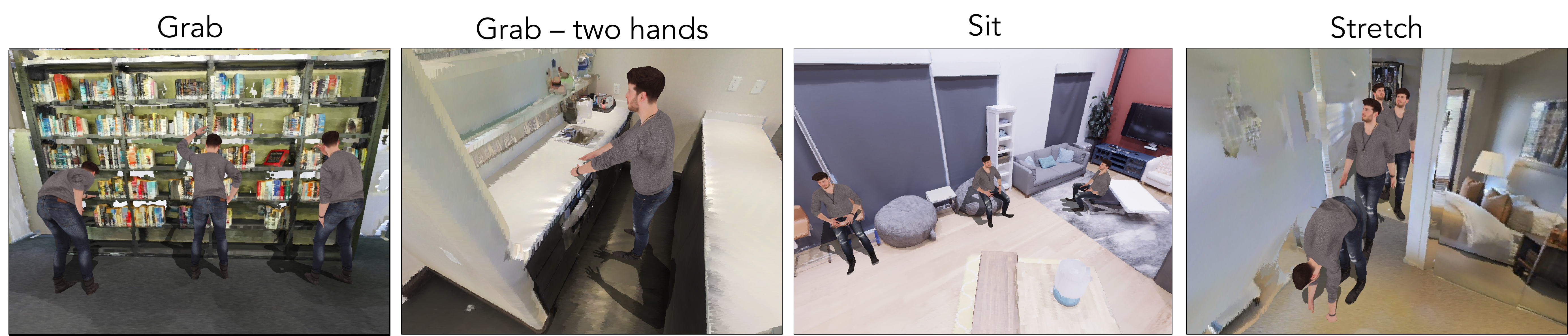}

\caption{The keypoint representation allows us to generate diverse and highly controllable motion. We show here examples of different grabbing, sitting and newly defined motions. }
\label{fig:qual_pose}
\end{figure*}
\end{center}

\vspace{-16mm}

\textbf{Can \textit{WalkNet} be replaced with other path following methods?}
We provide comparisons with SAMP, Wang et al, and GAMMA which all navigate A* paths. As our experiments illustrate, our method outperforms these existing methods for navigation. For further completeness, we trained 
the SoTA walking method, MoGlow \cite{henter2020moglow}, on our walking data . When deployed on 150-200 meter long A* paths, it produces significant foot-skating after about 30 secs. We hypothesize that this occurs because MoGlow synthesizes motion in an egocentric coordinate frame, and hence the control signal provided by A* changes rapidly which leads MoGlow to synthesize motion with significant drift. We compare our method to MoGlow on these paths using a user study with 36 participants in Tab.~\ref{tab:table}, where our approach outperforms MoGlow.

\textbf{How well does language based keypoint placement work?}
In this experiment, we compare motion synthesized using manual keypoint placement with language based keypoint placement. We synthesize 5 motion sequences using keypoints generated by these two approaches and compare the sythesized sequences using a user study with 36 participants.
When used for motion synthesis, these KPs produce similar quality as manual KP placement (Tab.~\ref{tab:table}).

\textbf{How long does it take for a user to provide keypoints manually?}
We develop a user interface which allows users to navigate 3D scenes and to click on locations of interaction. We instruct 7 participants how to navigate 3D scenes with our user interface. On average it takes 245 seconds for users to learn the interface. We then ask each user to provide 5 sets of 3 action keypoints (the location of the root and the two feet or the location of one hand and two feet) for a total of 15 keypoints per scene in 5 different 3D scenes. On average it takes 125 seconds to select  these points per scene.

\subsection{Qualitative Results}
\label{subsec:qual}
Please watch the supp. video for qualitative evaluation. %
In Figure~\ref{fig:qual_scene}, we demonstrate examples of motion generated in scenes from 4 different datasets: Replica~\cite{replica19arxiv}, Matterport\cite{niessner2017Matterport3D}, HPS\cite{mir20hps} and Scannet\cite{dai2017scannet}. Morover, representing the motion as Action Keypoints allows us to have high control and diversity over the generated motions. In Figure~\ref{fig:qual_pose} we show how this representation allows us to sit or pick objects at different heights (left column), or generate actions such as grabbing with two hands or stretching.

\section{Limitations and Conclusions}
\label{sec:conclusion}

We presented the first method to synthesize continual human motion in scenes from the HPS, Matterport, ScanNet, and Replica. Our core contribution is a novel method for long-range motion synthesis via iterative canonicalization and the use of keypoints to decouple scene reasoning from motion synthesis, and provide a flexible interface to synthesize motion. We demonstrated that our method works better than existing solutions that generate motion in 3D scenes.

While our approach presents an important step towards long-range motion synthesis in 3D scenes, it also has limitations: It assumes a horizontal floor and thus does not support scenes with uneven floors. It also assumes valid keypoint configurations: if the keypoints provided by the user do not conform to a valid pose, the pose produced by the IK step will not look realistic, producing unnatural motion. In the future we hope to remove this limitation by reducing the number of required keypoint inputs. We hope that the proposed approach drives new research towards continual human motion in arbitrary 3D scenes.

\section{Acknowledgements}
This work is funded by the Deutsche Forschungsgemeinschaft (DFG, German Research Foundation) - 409792180 (Emmy Noether Programme, project: Real Virtual Humans) and German Federal Ministry of Education and Research (BMBF): Tübingen AI Center, FKZ: 01IS18039A. Gerard Pons-Moll is a member of the Machine Learning Cluster of Excellence, EXC number 2064/1 – Project number 390727645. The project was made possible by funding from the Carl Zeiss Foundation.

\bibliographystyle{splncs04}
\bibliography{egbib}
\end{document}


\title{Supplementary for Back to the Origin: Generating Continual Human Motion in Any 3D Scene}

\maketitle

\section{Archtecture Details}
\label{sec:arch}
\begin{figure}
    \centering
    \includegraphics{}
    \caption{Caption}
    \label{fig:my_label}
\end{figure}

\section{IK Details}
\label{sec:arch}

\section{AMASS Data Normalization}
\label{sec:related}

\section{Walk Data Augmentation}
\label{sec:method}

\section{Path Planning}
\label{sec:planning}

\section{User Study Details}
\label{sec:planning}

\section{Different values of M}
\label{sec:planning}

\section{Comparison with SAMP}
\label{sec:planning}

\section{Comparison with Wang et al.}
\label{sec:planning}
\section{Placement of static walking poses}
\label{sec:static}

\bibliographystyle{splncs04}
\bibliography{egbib}